\documentclass[10pt,twocolumn,letterpaper]{article}
\usepackage[T1]{fontenc}
\usepackage{iccv}
\usepackage{times}

\usepackage[font={footnotesize}]{caption}
\usepackage[linesnumbered, ruled]{algorithm2e}
\usepackage[percent]{overpic}
\usepackage{amsmath}
\usepackage{amssymb}
\usepackage{booktabs}
\usepackage{color}
\usepackage{comment}
\usepackage{contour}
\usepackage{dsfont}
\usepackage{floatrow}
\usepackage{graphicx}
\usepackage{layouts}
\usepackage{listings}
\usepackage{multirow}
\usepackage{subcaption}
\usepackage{tabularx}
\usepackage{tcolorbox}
\usepackage{transparent}
\usepackage{url}
\usepackage{verbatim}
\usepackage{xspace}

\usepackage[pagebackref=true,breaklinks=true,letterpaper=true,colorlinks,bookmarks=false]{hyperref}
\usepackage{cleveref}

\iccvfinalcopy
\def\iccvPaperID{3193}
\ificcvfinal\fi

\newfloatcommand{capbtabbox}{table}[][\FBwidth]

\newcommand{\bx}{\mathbf{x}}
\newcommand{\by}{\mathbf{y}}

\newcommand{\methodname}{Invariant Information Clustering\xspace}
\newcommand{\methodnameshort}{IIC\xspace}

\newcommand{\cmt}[1]{\ignorespaces}

\newcommand{\matP}{\mathbf{P}}  

\DeclareFixedFont{\ttb}{T1}{txtt}{bx}{n}{6.5} 
\DeclareFixedFont{\ttm}{T1}{txtt}{m}{n}{6.5}  

\vbadness=\maxdimen

\makeatletter
\renewcommand{\paragraph}{%
  \@startsection{paragraph}{4}%
  {\z@}{0.5em}{-1em}%
  {\normalfont\normalsize\bfseries}%
}
\makeatother

\title{\methodname for \\ Unsupervised Image Classification and Segmentation}
\author{Xu Ji\\
University of Oxford\\
{\tt\small xuji@robots.ox.ac.uk}
\and
Jo\~ao F. Henriques\\
University of Oxford\\
{\tt\small joao@robots.ox.ac.uk}
\and
Andrea Vedaldi\\
University of Oxford\\
{\tt\small vedaldi@robots.ox.ac.uk}}

\begin{document}
\maketitle
\begin{abstract}

We present a novel clustering objective that learns a neural network classifier from scratch, given only unlabelled data samples.
The model discovers clusters that accurately match semantic classes, achieving state-of-the-art results in eight unsupervised clustering benchmarks spanning image classification and segmentation.
These include STL10, an unsupervised variant of ImageNet, and CIFAR10, where we significantly beat the accuracy of our closest competitors by 6.6 and 9.5 absolute percentage points respectively.
The method is not specialised to computer vision and operates on any paired dataset samples; in our experiments we use random transforms to obtain a pair from each image. 
The trained network directly outputs semantic labels, rather than high dimensional representations that need external processing to be usable for semantic clustering.
The objective is simply to maximise mutual information between the class assignments of each pair.
It is easy to implement and rigorously grounded in information theory, meaning we effortlessly avoid degenerate solutions that other clustering methods are susceptible to.
In addition to the fully unsupervised mode, we also test two semi-supervised settings. The first achieves 88.8\% accuracy on STL10 classification, setting a new global state-of-the-art over all existing methods (whether supervised, semi-supervised or unsupervised).
The second shows robustness to 90\% reductions in label coverage, of relevance to applications that wish to make use of small amounts of labels. \lstinline[basicstyle=\ttfamily]{github.com/xu-ji/IIC} 
\end{abstract}

\section{Introduction}\label{s:intro}

Most supervised deep learning methods require large quantities of manually labelled data, limiting their applicability in many scenarios.
This is true for large-scale image classification and even more for segmentation (pixel-wise classification) where the annotation cost per image is very high~\cite{lin2014microsoft,girshick2014rich}.
Unsupervised clustering, on the other hand, aims to group data points into classes entirely without labels~\cite{hartigan1972direct}.
Many authors have sought to combine mature clustering algorithms with deep learning, for example by bootstrapping network training with k-means style objectives~\cite{xie2016unsupervised, haeusser2018associative, caron2018deep}.
However, trivially combining clustering and representation learning methods often leads to degenerate solutions~\cite{caron2018deep, xie2016unsupervised}.
It is precisely to prevent such degeneracy that cumbersome pipelines --- involving pre-training, feature post-processing (whitening or PCA), clustering mechanisms external to the network --- have evolved~\cite{caron2018deep,doersch2015unsupervised,dosovitskiy2015discriminative,xie2016unsupervised}.

\begin{figure}[t]

\includegraphics[height=1.1\textwidth]{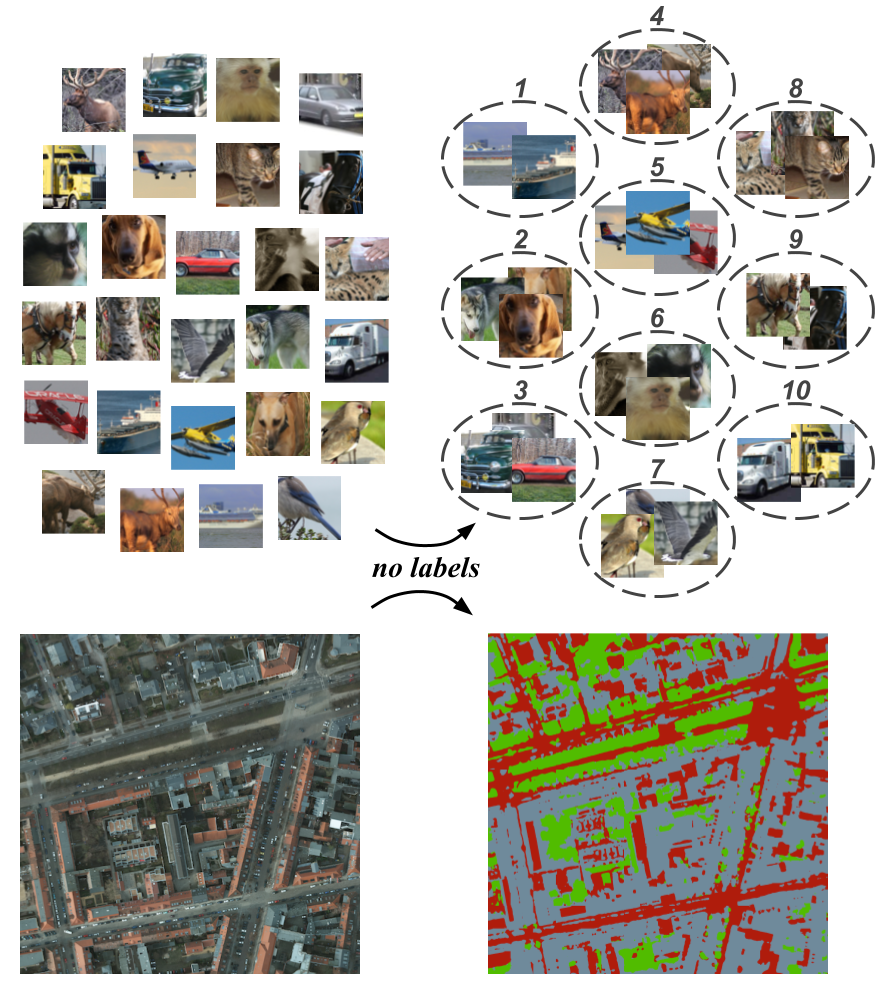}

\caption{\label{f:splash} Models trained with \methodnameshort on entirely unlabelled data learn to cluster images (top, STL10) and patches (bottom, Potsdam-3). The raw clusters found directly correspond to semantic classes (dogs, cats, trucks, roads, vegetation etc.) with state-of-the-art accuracy.  Training is end-to-end and randomly initialised, with no heuristics used at any stage.}
\end{figure}

In this paper, we introduce \methodname (\methodnameshort), a method that addresses this issue in a more principled manner.
\methodnameshort is a generic clustering algorithm that directly trains a randomly initialised neural network into a classification function, end-to-end and without any labels.
It involves a simple objective function, which is the mutual information between the function's classifications for paired data samples. The input data can be of any modality and, since the clustering space is discrete, mutual information can be computed exactly.

\begin{figure*}
\setlength\tabcolsep{2.2pt} 

\begin{tabular}{c c c c c c}
\includegraphics[height=0.16\textwidth]{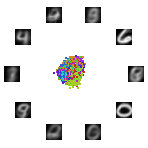} & 
\includegraphics[height=0.16\textwidth]{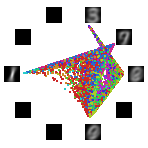} & 
\includegraphics[height=0.16\textwidth]{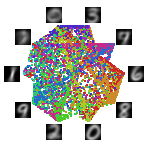} & 
\includegraphics[height=0.16\textwidth]{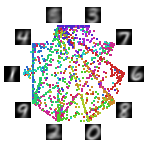} & 
\includegraphics[height=0.16\textwidth]{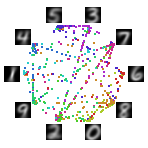} & 
\includegraphics[height=0.16\textwidth]{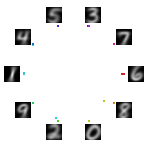} 
\end{tabular}

\caption{\label{f:mnist_dots} Training with \methodnameshort on unlabelled MNIST in successive epochs from random initialisation (left). The network directly outputs cluster assignment probabilities for input images, and each is rendered as a coordinate by convex combination of 10 cluster vertices. There is no cherry-picking as the entire dataset is shown in every snapshot. Ground truth labelling (unseen by model) is given by colour. At each cluster the average image of its assignees is shown. With neither labels nor heuristics, the clusters discovered by \methodnameshort correspond perfectly to unique digits, with one-hot certain prediction (right).}
\end{figure*}

Despite its simplicity, \methodnameshort is intrinsically robust to two issues that affect other methods.
The first is clustering degeneracy, which is the tendency for a single cluster to dominate the predictions or for clusters to disappear (which can be observed with k-means, especially when combined with representation learning~\cite{caron2018deep}).
Due to the entropy maximisation component within mutual information, the loss is not minimised if all images are assigned to the same class. At the same time, it is optimal for the model to predict for each image a single class with certainty (i.e. one-hot) due to the conditional entropy minimisation~(\cref{f:mnist_dots}). The second issue is noisy data with unknown or distractor classes (present in STL10~\cite{coates2011analysis} for example).
\methodnameshort addresses this issue by employing an auxiliary output layer that is parallel to the main output layer, trained to produce an overclustering (i.e.\ same loss function but greater number of clusters than the ground truth) that is ignored at test time.
Auxiliary overclustering is a general technique that could be useful for other algorithms.
These two features of \methodnameshort contribute to making it the only method amongst our unsupervised baselines that is robust enough to make use of the noisy unlabelled subset of STL10, a version of ImageNet~\cite{deng2009imagenet} specifically designed as a benchmark for unsupervised clustering.

In the rest of the paper, we begin by explaining the difference between semantic clustering and intermediate representation learning~(\cref{s:related}), which separates our method from the majority of work in unsupervised deep learning. 
We then describe the theoretical foundations of \methodnameshort in statistical learning~(\cref{s:method}), demonstrating that maximising mutual information between pairs of samples under a bottleneck is a principled clustering objective which is equivalent to distilling their shared abstract content (co-clustering).
We propose that for static images, an easy way to generate pairs with shared abstract content from unlabelled data is to take each image and its random transformation, or each patch and a neighbour.
We show that maximising MI automatically avoids degenerate solutions and can be written as a convolution in the case of segmentation, allowing for efficient implementation with any deep learning library.

We perform experiments on a large number of datasets~(\cref{s:experiments}) including STL, CIFAR, MNIST, COCO-Stuff and Potsdam, setting a new state-of-the-art on unsupervised clustering and segmentation in all cases, with results of 59.6\%, 61.7\% and 72.3\% on STL10, CIFAR10 and COCO-Stuff-3 beating the closest competitors (53.0\%, 52.2\%, 54.0\%) with significant margins.
Note that training deep neural networks to perform large scale, real-world segmentations from scratch, without labels or heuristics, is a highly challenging task with negligible precedent.
We also perform an ablation study and additionally test two semi-supervised modes, setting a new global state-of-the-art of 88.8\% on STL10 over all supervised, semi-supervised and unsupervised methods, and demonstrating the robustness in semi-supervised accuracy when 90\% of labels are removed.

\section{Related work}\label{s:related}

\begin{figure}[t]
\centering
\includegraphics[width=0.95\columnwidth]{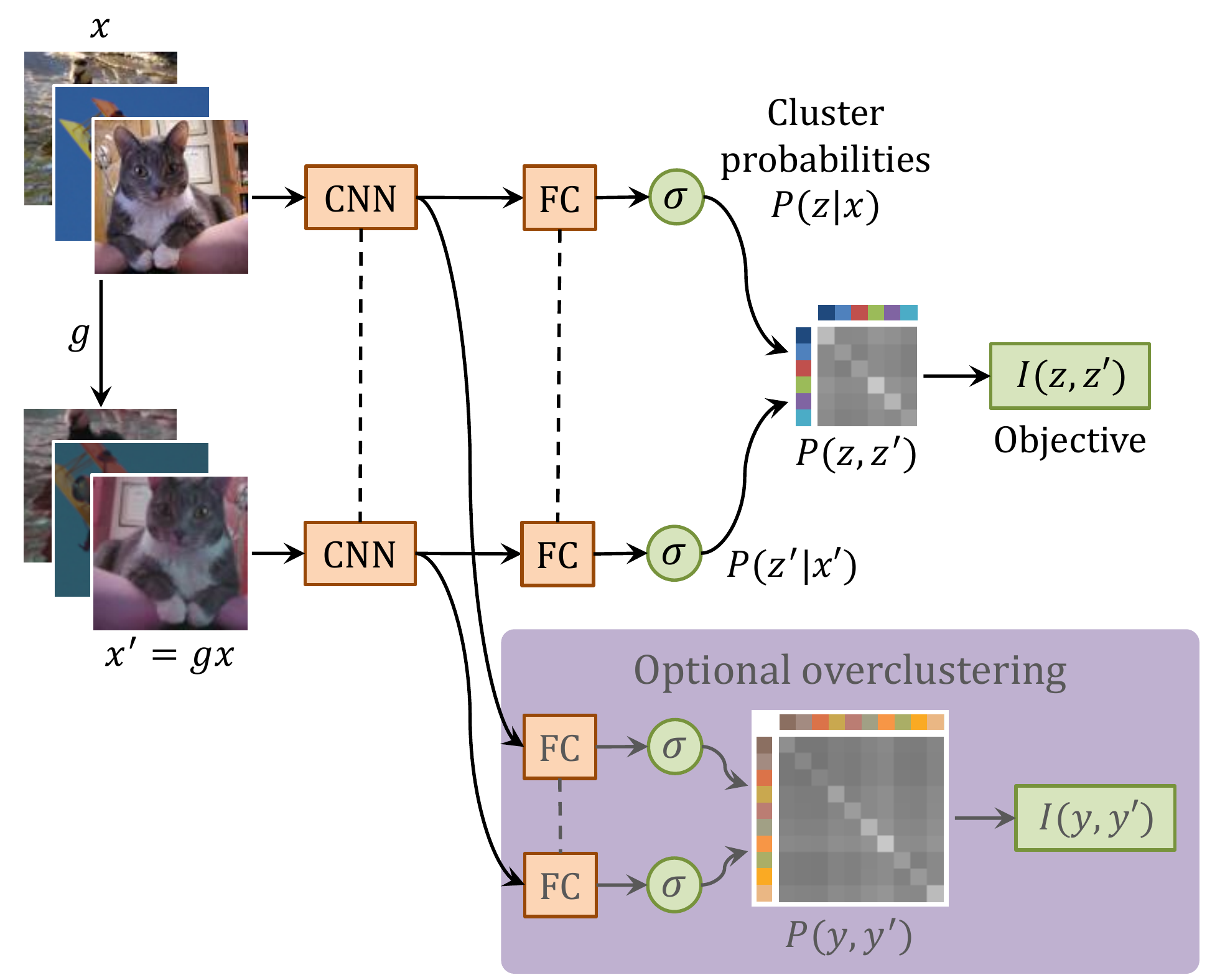}
\caption{\label{f:overview}\methodnameshort for image clustering. Dashed line denotes shared parameters, $g$ is a random transformation, and $I$ denotes mutual information~(\cref{e:loss_expanded}).}
\end{figure}

\paragraph{Co-clustering and mutual information.}

The use of information as a criterion to learn representations is not new. One of the earliest works to do so is by Becker and Hinton~\cite{becker1992self}.
More generally, learning from paired data has been explored in co-clustering~\cite{hartigan1972direct, dhillon2003information} and in other works~\cite{wang2010information} that build on the information bottleneck principle~\cite{friedman2001multivariate}.

Several recent papers have used information as a tool to train deep networks in particular.
IMSAT~\cite{hu2017learning} maximises mutual information between data and its representation and DeepINFOMAX~\cite{hjelm2018learning} maximizes information between spatially-preserved features and compact features.
However, IMSAT and DeepINFOMAX combine information with other criteria, whereas in our method information is the only criterion used.
Furthermore, both IMSAT and DeepINFOMAX compute mutual information over continuous random variables, which requires complex estimators~\cite{belghazi2018mine}, whereas \methodnameshort does so for discrete variables with simple and exact computations.
Finally, DeepINFOMAX considers the information $I(\bx, f(\bx))$ between the features $\bx$ and a deterministic function $f(\bx)$ of it, which is in principle the same as the entropy $H(\bx)$; in contrast, in \methodnameshort information does not trivially reduce to  entropy.

\paragraph{Semantic clustering versus intermediate representation learning.}
In semantic clustering, the learned function directly outputs discrete assignments for high level (i.e. semantic) clusters. Intermediate representation learners, on the other hand, produce continuous, distributed, high-dimensional representations that must be post-processed, for example by k-means, to obtain the discrete low-cardinality assignments required for unsupervised semantic clustering. The latter includes objectives such as generative autoencoder image reconstruction~\cite{vincent2010stacked},  triplets~\cite{schultz2004learning} and spatial-temporal order or context prediction~\cite{lee2017unsupervised,cruz2017deeppermnet,doersch2015unsupervised}, for example predicting patch proximity~\cite{isola2015learning}, solving jigsaw puzzles~\cite{noroozi2016unsupervised} and inpainting~\cite{pathak2016context}. Note it also includes a number of clustering methods (DeepCluster~\cite{caron2018deep}, exemplars~\cite{dosovitskiy2015discriminative}) where the clustering is only auxiliary; a clustering-style objective is used but does not produce groups with semantic correspondence. For example, DeepCluster~\cite{caron2018deep} is a state-of-the-art method for learning highly-transferable intermediate features using overclustering as a proxy task, but does not automatically find semantically meaningful clusters. As these methods use auxiliary objectives divorced from the semantic clustering objective, it is unsurprising that they perform worse than \methodnameshort~(\cref{s:experiments}), which directly optimises for it, training the network end-to-end with the final clusterer implicitly wrapped inside.

\paragraph{Optimising image-to-image distance.}

Many approaches to deep clustering, whether semantic or auxiliary, utilise a distance function between input images that approximates a given grouping criterion.
Agglomerative clustering~\cite{bautista2016cliquecnn} and partially ordered sets~\cite{bautista2017deep} of HOG features~\cite{dalal2005histograms} have been used to group images, and exemplars~\cite{dosovitskiy2015discriminative} define a group as a set of random transformations applied to a single image. Note the latter does not scale easily, in particular to image segmentation where a single $200\times 200$ image would call for 40k classes. DAC~\cite{chang2017deep}, JULE~\cite{yang2016joint}, DeepCluster~\cite{caron2018deep}, ADC~\cite{haeusser2018associative} and DEC~\cite{xie2016unsupervised} rely on the inherent visual consistency and disentangling properties~\cite{greff2015binding} of CNNs to produce cluster assignments, which are processed and reinforced in each iteration. 
The latter three are based on k-means style mechanisms to refine feature centroids, which is prone to degenerate solutions~\cite{caron2018deep} and thus needs explicit prevention mechanisms such as pre-training, cluster-reassignment or feature cleaning via PCA and whitening~\cite{xie2016unsupervised, caron2018deep}.

\paragraph{Invariance as a training objective.}

Optimising for function outputs to be persistent through spatio-temporal or non-material distortion is an idea shared by \methodnameshort with several works, including exemplars~\cite{dosovitskiy2015discriminative}, IMSAT~\cite{hu2017learning}, proximity prediction~\cite{isola2015learning}, the denoising objective of Tagger~\cite{greff2016tagger}, temporal slowness constraints~\cite{zou2012deep}, and optimising for features to be invariant to local image transformations~\cite{sohn2012learning,hui2013direct}.
More broadly, the problem of modelling data transformation has received significant attention in deep learning, one example being the transforming autoencoder~\cite{hinton2011transforming}.

\section{Method}\label{s:method}

First we introduce a generic objective, \methodname, which can be used to cluster any kind of unlabelled paired data by training a network to predict cluster identities~(\cref{s:generic_clustering}).
We then apply it to image clustering~(\cref{s:image_clustering}, \cref{f:overview} and \cref{f:mnist_dots}) and segmentation~(\cref{s:image_segmentation}), by generating the required paired data using random transformations and spatial proximity. 


\subsection{\methodname}\label{s:generic_clustering}

Let $\bx,\bx' \in \mathcal{X}$ be a paired data sample from a joint probability distribution $P(\bx,\bx')$.
For example, $\bx$ and $\bx'$ could be different images containing the same object.
The goal of \methodname (\methodnameshort) is to learn a representation $\Phi:\mathcal{X}\rightarrow\mathcal{Y}$ that preserves what is in common between $\bx$ and $\bx'$ while discarding instance-specific details.
The former can be achieved by maximizing the mutual information between encoded variables:
\begin{equation}\label{e:info_orig}
\max_\Phi I(\Phi(\bx),\Phi(\bx')),
\end{equation}
which is equivalent to maximising the predictability of $\Phi(\bx)$ from $\Phi(\bx')$ and vice versa. 

An effect of equation~\cref{e:info_orig}, in general, is to make representations of paired samples the same.
However, it is not the same as merely minimising representation distance, as done for example in methods based on k-means~\cite{caron2018deep,haeusser2018associative}: the presence of entropy within $I$ allows us to avoid degeneracy, as discussed in detail below.



If $\Phi$ is a neural network with a small output capacity (often called a ``bottleneck''), \cref{e:info_orig} also has the effect of discarding instance-specific details from the data.
Clustering imposes a natural bottleneck, since the representation space is $\mathcal{Y}=\{1,\dots,C\}$, a finite set of class indices (as opposed to an infinite vector space). 
Without a bottleneck, i.e. assuming unbounded capacity, \cref{e:info_orig} is trivially solved by setting $\Phi$ to the identity function because of the data processing inequality~\cite{cover2012elements}, i.e. $I(\bx, \bx') \geq I(\Phi(\bx), \Phi(\bx'))$.

\newcommand{\bPhi}{\Phi}

Since our goal is to learn the representation with a deep neural network, we consider soft rather than hard clustering, meaning the neural network $\bPhi$ is terminated by a (differentiable) softmax layer.
Then the output $\bPhi(\bx) \in [0,1]^C$ can be interpreted as the distribution of a discrete random variable $z$ over $C$ classes, formally given by $P(z = c|\bx) = \bPhi_c(\bx)$. Making the output probabilistic amounts to allowing for uncertainty in the cluster assigned to an input.

Consider now a pair of such cluster assignment variables $z$ and $z'$ for two inputs $\bx$ and $\bx'$ respectively.
Their conditional joint distribution is given by
$
P(z=c,z'=c'|\bx,\bx')
=  \bPhi_c(\bx) \cdot \bPhi_{c'}(\bx').
$
This equation states that $z$ and $z'$ are independent
when conditioned on specific inputs $\bx$ and $\bx'$; however, in general they are \emph{not} independent after marginalization over a dataset of input pairs $(\bx_i,\bx'_i)$, $i=1,\dots,n$.
For example, for a trained classification network $\Phi$ and a dataset of image pairs where each image contains the same object of its pair but in a randomly different position, the random variable constituted by the class of the first of each pair, $z$, will have a strong statistical relationship with the random variable for the class of the second of each pair, $z'$; one is predictive of the other (in fact identical to it, in this case) so they are highly dependent. 
After marginalization over the dataset (or batch, in practice), the joint probability distribution is given by the $C\times C$ matrix $\matP$, where each element at row $c$ and column $c'$ constitutes $\matP_{cc'} = P(z=c,z'=c')$:
\begin{equation}\label{e:joint}
\matP
=
\frac{1}{n} 
\sum_{i=1}^n \bPhi(\bx_i) \cdot\bPhi(\bx'_i)^\top.
\end{equation}
The marginals $\matP_c=P(z=c)$ and $\matP_{c'}=P(z'=c')$ can be obtained by summing over the rows and columns of this matrix. 
As we generally consider symmetric problems, where for each $(\bx_i,\bx'_i)$ we also have $(\bx'_i,\bx_i)$, $\matP$ is symmetrized using $(\matP+\matP^\top)/2$.

Now the objective function~\cref{e:info_orig} can be computed by plugging the matrix $\matP$ into the expression for mutual information~\cite{learned2013entropy}, which results in the formula:
\begin{equation}\label{e:loss_expanded}
I(z,z') =
I(\matP) =
\sum_{c=1}^C
\sum_{c'=1}^C
\matP_{cc'} \cdot 
\ln\frac{\matP_{cc'}}{\matP_c \cdot \matP_{c'}}.
\end{equation}


  
\paragraph{Why degenerate solutions are avoided.}\label{s:equalization}

Mutual information~\eqref{e:loss_expanded} expands to
$
I(z,z')
=
H(z) - H(z|z')
$.
Hence, maximizing this quantity trades-off minimizing the conditional cluster assignment entropy $H(z|z')$ and maximising individual cluster assignments entropy $H(z)$.
The smallest value of $H(z|z')$ is 0, obtained when the cluster assignments are exactly predictable from each other.
The largest value of $H(z)$ is $\ln C$, obtained when all clusters are equally likely to be picked. This occurs when the data is assigned evenly between the clusters, equalizing their mass. Therefore the loss is not minimised if all samples are assigned to a single cluster (i.e. output class is identical for all samples).
Thus as maximising mutual information naturally balances reinforcement of predictions with mass equalization, it avoids the tendency for degenerate solutions that algorithms which combine k-means with representation learning are susceptible to~\cite{caron2018deep}.
For further discussion of entropy maximisation, and optionally how to prioritise it with an entropy coefficient, see supplementary material.

\paragraph{Meaning of mutual information.}

The reader may now wonder what are the benefits of maximising mutual information, as opposed to merely maximising entropy.
Firstly, due to the soft clustering, entropy alone could be maximised trivially by setting all prediction vectors $\Phi(\bx)$ to uniform distributions, resulting in no clustering.
This is corrected by the conditional entropy component, which encourages deterministic one-hot predictions.
For example, even for the degenerate case of identical pairs $\bx = \bx'$, the \methodnameshort objective encourages a deterministic clustering function (i.e.~$\Phi(\bx)$ is a one-hot vector) as this results in null conditional entropy $H(z|z')=0$. 
Secondly, the objective of \methodnameshort is to find what is common between two data points that share redundancy, such as different images of the same object, explicitly encouraging distillation of the common part while ignoring the rest, i.e.~instance details specific to one of the samples. This would not be possible without pairing samples.

\subsection{Image clustering}\label{s:image_clustering}

\methodnameshort requires a source of paired samples $(\bx,\bx')$, which are often unavailable in unsupervised image clustering applications. In this case, we propose to use \emph{generated} image pairs, consisting of image $\bx$ and its \emph{randomly perturbed} version $\bx' = g\bx$. The objective~\cref{e:info_orig} can thus be written as:
\begin{equation}\label{e:info_clustering}
\max_\Phi I(\Phi(\bx),\Phi(g\bx)),
\end{equation}
where both image $\bx$ and transformation $g$ are random variables.
Useful $g$ could include scaling, skewing, rotation or flipping (geometric), changing contrast and colour saturation (photometric), or any other perturbation that is likely to leave the content of the image intact.
\methodnameshort can then be used to recover the factor which is \emph{invariant} to which of the pair is picked. 
The effect is to learn a function that partitions the data such that clusters are closed to the perturbations, without dropping clusters. 
The objective is simple enough to be written in six lines of PyTorch code ~(\cref{f:code}).

 \definecolor{codegreen}{rgb}{0,0.6,0}
\definecolor{codegray}{rgb}{0.5,0.5,0.5}
\definecolor{codepurple}{rgb}{0.58,0,0.82}
\definecolor{backcolour}{rgb}{0.95,0.95,0.92}

\lstdefinelanguage{mypython}{morekeywords={def, return}}

\lstdefinestyle{mystyle}{
    language=mypython,
    commentstyle=\color{codegreen},
    keywordstyle=\color{codegreen},
    numberstyle=\color{codegray},
    stringstyle=\color{magenta},
    basicstyle=\ttfamily\fontsize{6.6pt}{7.92pt}\selectfont,
    breakatwhitespace=true,         
    breaklines=true,                 
    captionpos=b,                    
    keepspaces=true,                 
    numbers=none,                    
    showspaces=false,                
    showstringspaces=false,
    showtabs=false,                  
    tabsize=2,
    frame=single,
}

\begin{figure}[t]

\begin{lstlisting}[style=mystyle]
def IIC(z, zt, C=10):
  P = (z.unsqueeze(2) * zt.unsqueeze(1)).sum(dim=0)
  P = ((P + P.t()) / 2) / P.sum()
  P[(P < EPS).data] = EPS
  Pi = P.sum(dim=1).view(C, 1).expand(C, C)
  Pj = P.sum(dim=0).view(1, C).expand(C, C)
  return (P * (log(Pi) + log(Pj) - log(P))).sum()
\end{lstlisting}

\caption{\label{f:code} \methodnameshort objective in PyTorch. Inputs \texttt{z} and \texttt{zt} are $n \times C$ matrices, with $C$ predicted cluster probabilities for $n$ sampled pairs (i.e. CNN softmaxed predictions). For example, the prediction for each image in a dataset and its transformed version (e.g. using standard data augmentation).}
\end{figure}

\paragraph{Auxiliary overclustering.}\label{s:overcluster}

For certain datasets (e.g. STL10), training data comes in two types: one known to contain only relevant classes and the other known to contain irrelevant or distractor classes. It is desirable to train a clusterer specialised for the relevant classes, that still benefits from the context provided by the distractor classes, since the latter is often much larger (for example 100K compared to 13K for STL10).
Our solution is to add an auxiliary overclustering head to the network~(\cref{f:overview}) that is trained with the full dataset, whilst the main output head is trained with the subset containing only relevant classes.
This allows us to make use of the noisy unlabelled subset despite being an unsupervised clustering method. 
Other methods are generally not robust enough to do so and thus avoid the 100k-samples unlabelled subset of STL10 when training for unsupervised clustering~(\cite{chang2017deep, haeusser2018associative,xie2016unsupervised}).
Since the auxiliary overclustering head outputs predictions over a larger number of clusters than the ground truth, whilst still maintaining a predictor that is matched to ground truth number of clusters (the main head), it can be useful in general for increasing expressivity in the learned feature representation, even for datasets where there are no distractor classes~\cite{caron2018deep}.

\subsection{Image segmentation}\label{s:image_segmentation}

\methodnameshort can be applied to image segmentation identically to image clustering, except for two modifications.
Firstly, since predictions are made for each pixel densely, clustering is applied to image patches (defined by the receptive field of the neural network for each output pixel) rather than whole images.
Secondly, unlike with whole images, one has access to the spatial relationships between patches.
Thus, we can add \emph{local spatial invariance} to the list of geometric and photometric invariances in~\cref{s:image_clustering}, meaning we form pairs of patches not only via synthetic perturbations, but also by extracting pairs of adjacent patches in the image.



In detail, let the RGB image $\bx\in\mathbb{R}^{3\times H\times W}$ be a tensor, $u \in \Omega = \{1,\dots,H\}\times\{1,\dots,W\}$ a pixel location, and $\bx_u$ a patch centered at $u$.
We can form a pair of patches $(\bx_u,\bx_{u+t})$ by looking at location $u$ and its neighbour $u+t$ at some small displacement $t\in T \subset\mathbb{Z}^2$.
The cluster probability vectors for all patches $\bx_u$ can be read off as the column vectors $\Phi(\bx_u) = \Phi_u(\bx) \in [0,1]^C$ of the tensor $\Phi(\bx)\in [0,1]^{C\times H\times W}$, computed by a single application of the convolutional network $\Phi$.
Then, to apply \methodnameshort, one simply substitutes pairs $(\Phi_u(\bx)$, $\Phi_{u + t}(\bx))$ in the calculation of the joint probability matrix~\eqref{e:joint}.




The geometric and photometric perturbations used before for whole image clustering can be applied to individual patches too.
Rather than transforming patches individually, however, it is much more efficient to transform all of them in parallel by perturbing the entire image.
Any number or combination of these invariances can be chained and learned simultaneously; the only detail is to ensure indices of the original image and transformed image class probability tensors line up, meaning that predictions from patches which are intended to be paired together do so.

Formally, if the image transformation $g$ is a geometric transformation, the vector of cluster probabilities $\Phi_u(\bx)$ will not correspond to $\Phi_u(g\bx)$; rather, it will correspond to $\Phi_{g(u)}(g\bx)$ because patch $\bx_u$ is sent to patch $\bx_{g(u)}$ by the transformation.
All vectors can be paired at once by applying the reverse transformation $g^{-1}$ to the tensor $\Phi(g\bx)$, as
$
[g^{-1}\Phi(g\bx)]_u = \Phi_{g(u)}(g\bx).
$
For example, flipping the input image will require flipping the resulting probability tensor back. In general, the perturbation $g$ can incorporate geometric and photometric transformations, and $g^{-1}$ only needs to undo geometric ones. The segmentation objective is thus:
\begin{flalign}\label{e:info_seg}
\max_{\Phi}&
\frac{1}{|T|}\sum_{t \in T} I(\matP_t),
\\[-1.5em]
\matP_t &=
\frac{1}{n|G||\Omega|}
\sum_{i=1}^n \sum_{g\in G}
\overbrace{
  \sum_{u\in \Omega}
 \Phi_{u}(\bx_i) \cdot [g^{-1}\Phi(g\bx_i)]_{u+t}^\top.
}^{\text{Convolution}}\nonumber
\end{flalign}
Hence the goal is to maximize the information between each patch label $\Phi_{u}(\bx_i)$  and the patch label $[g^{-1}\Phi(g\bx_i)]_{u+t}$ of its transformed neighbour patch, in expectation over images $i=1,\dots,n$, patches $u\in\Omega$ within each image, and perturbations $g\in G$.
Information is in turn averaged over all neighbour displacements $t\in T$ (which was found to perform slightly better than averaging over $t$ before computing information; see supplementary material).

\paragraph{Implementation.}
The joint distribution of~\cref{e:info_seg} for all displacements $t \in T$ can be computed in a simple and highly efficient way.
Given two network outputs for one batch of image pairs $\by = \Phi(\bx), \by' = \Phi(g\bx)$ where $\by, \by' \in\mathbb{R}^{n\times C\times H\times W}$, we first bring $\by'$ back into the coordinate-space of $\by$ by using a bilinear resampler\footnote{The core differentiable operator in spatial transformer networks~\cite{jaderberg2015spatial}.}~\cite{jaderberg2015spatial}, which inverts any geometrical transforms in $g$, $\by' \leftarrow g^{-1}\by'$.
Then, the inner summation in \cref{e:info_seg} reduces to the convolution of the two tensors. Using any standard deep learning framework, this can be achieved by swapping the first two dimensions of each of $y$ and $y'$, computing $\matP = y \ast y'$ (a 2D convolution with padding $d$ in both dimensions), and normalising the result to produce $\matP \in [0,1]^{C\times C\times (2d+1)\times (2d+1)}$.

\section{Experiments}\label{s:experiments}

\begin{table}[t]
\setlength{\tabcolsep}{2.5pt}
\fontsize{8}{9}\selectfont 
\begin{tabular}{lcccc}
\toprule
& STL10 & CIFAR10 & CFR100-20 & MNIST \\
\midrule
Random network & 13.5 & 13.1 & 5.93 & 26.1 \\
K-means~\cite{zelnik2005self}$\dagger$  & 19.2 & 22.9 & 13.0 & 57.2 \\
Spectral clustering~\cite{wang2015optimized} & 15.9 & 24.7 & 13.6 & 69.6\\
Triplets~\cite{schultz2004learning}$\ddagger$ & 24.4 & 20.5 & 9.94 & 52.5 \\
AE~\cite{bengio2007greedy}$\ddagger$ & 30.3 & 31.4 & 16.5 & 81.2 \\
Sparse AE~\cite{ng2011sparse}$\ddagger$ & 32.0 & 29.7 & 15.7 & 82.7\\
Denoising AE~\cite{vincent2010stacked}$\ddagger$ & 30.2 & 29.7 & 15.1 & 83.2\\
Variational Bayes AE~\cite{kingma2013auto}$\ddagger$ & 28.2 & 29.1 & 15.2 & 83.2\\
SWWAE 2015~\cite{zhao2015stacked}$\ddagger$ & 27.0 & 28.4 & 14.7 & 82.5 \\
GAN 2015~\cite{radford2015unsupervised}$\ddagger$ & 29.8 & 31.5 & 15.1 & 82.8\\
JULE 2016~\cite{yang2016joint} & 27.7 & 27.2 & 13.7 & 96.4 \\
DEC 2016~\cite{xie2016unsupervised}$\dagger$ & 35.9 & 30.1 & 18.5 & 84.3 \\
DAC 2017~\cite{chang2017deep} & 47.0 & 52.2 & 23.8 & 97.8 \\
DeepCluster 2018~\cite{caron2018deep}$\dagger$ $\ddagger$ & 33.4$\star$ & 37.4$\star$ & 18.9$\star$ & 65.6 $\star$ \\
ADC 2018~\cite{haeusser2018associative} & 53.0 & 32.5 & 16.0$\star$ & 99.2 \\
\midrule
  \methodnameshort (lowest loss sub-head) & \textbf{59.6} \cmt{569} & \textbf{61.7} \cmt{640} &\textbf{25.7} \cmt{579} & \textbf{99.2} \cmt{685} \\
\methodnameshort (avg sub-head $\pm$ STD) & 59.8 & 57.6 & 25.5 & 98.4 \\[-0.2em]
  & ~~\scriptsize$\pm$ 0.844 & \scriptsize$\pm$ 5.01 & \scriptsize$\pm$ 0.462 & \scriptsize$\pm$ 0.652 \\
\bottomrule
\end{tabular}
\caption{\textbf{Unsupervised image clustering.} Legend: $\dagger$Method based on k-means. $\ddagger$Method that does not directly learn a clustering function and requires further application of k-means to be used for image clustering. $\star$Results obtained using our experiments with authors' original code.}\label{t:img_clus_iid}
\end{table}

\begin{table}[t]
\footnotesize    
\begin{tabular}{lc}
\toprule
& STL10 \\
\midrule
No auxiliary overclustering & 43.8 \cmt{692}\\
Single sub-head ($h=1$) & 57.6\cmt{693}\\
No sample repeats ($r=1$) & 47.0\cmt{694}\\
Unlabelled data segment ignored & 49.9\cmt{695}\\
\midrule
Full setting & \textbf{59.6} \cmt{569}\\
\bottomrule
\end{tabular}
\caption{\textbf{Ablations of \methodnameshort (unsupervised setting).} Each row shows a single change from the full setting. The full setting has auxiliary overclustering, 5 initialisation heads, 5 sample repeats, and uses the unlabelled data subset of STL10.}
\label{t:iid_imgclus_ablation}
\end{table}


We apply \methodnameshort to fully unsupervised image clustering and segmentation, as well as two semi-supervised settings. Existing baselines are outperformed in all cases.
We also conduct an analysis of our method via ablation studies. For minor details see supplementary material.

\subsection{Image clustering}\label{s:exp_img_clus}
\begin{figure*}[t]
\begin{center}
\setlength\tabcolsep{1.5pt} 
\begin{tabular}{c c c c c c c c c c c c c c c c}
\multicolumn{2}{c}{\textbf{\textit{Cat}}} &  
\multicolumn{2}{c}{\textbf{\textit{Dog}}} &
\multicolumn{2}{c}{\textbf{\textit{Bird}}} &
\multicolumn{2}{c}{\textbf{\textit{Deer}}} &
\multicolumn{2}{c}{\textbf{\textit{Monkey}}} &
\multicolumn{2}{c}{\textbf{\textit{Car}}} &
\multicolumn{2}{c}{\textbf{\textit{Plane}}} &
\multicolumn{2}{c}{\textbf{\textit{Truck}}} \\

\includegraphics[width=0.055\textwidth]{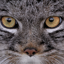} &
\includegraphics[width=0.055\textwidth]{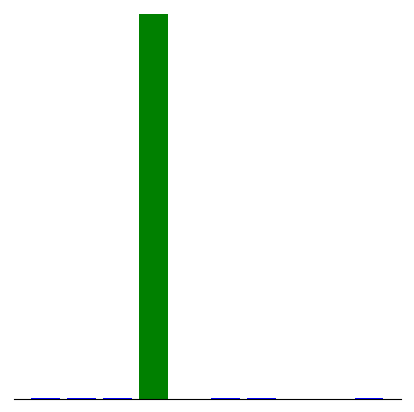} &

\includegraphics[width=0.055\textwidth]{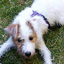} &
\includegraphics[width=0.055\textwidth]{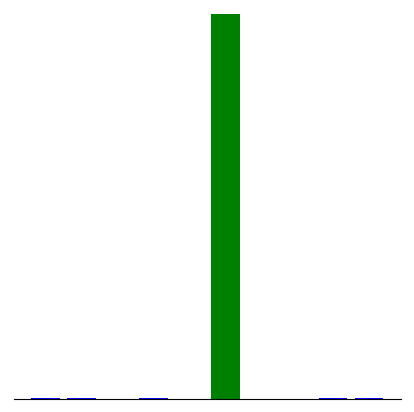} &

\includegraphics[width=0.055\textwidth]{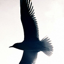} &
\includegraphics[width=0.055\textwidth]{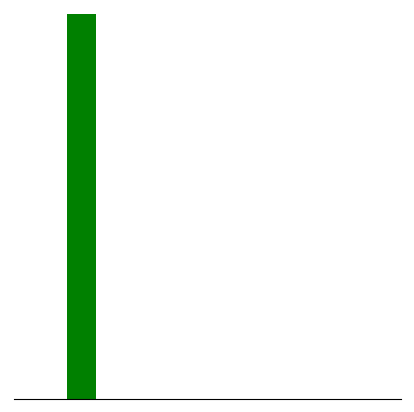} &

\includegraphics[width=0.055\textwidth]{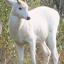} &
\includegraphics[width=0.055\textwidth]{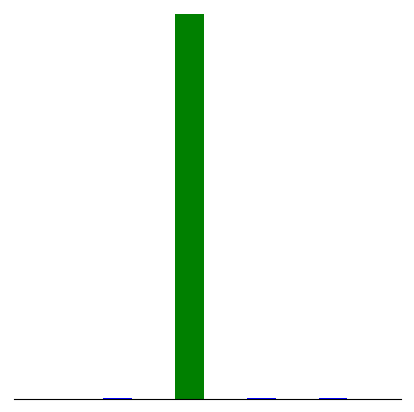} &

\includegraphics[width=0.055\textwidth]{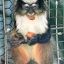} &
\includegraphics[width=0.055\textwidth]{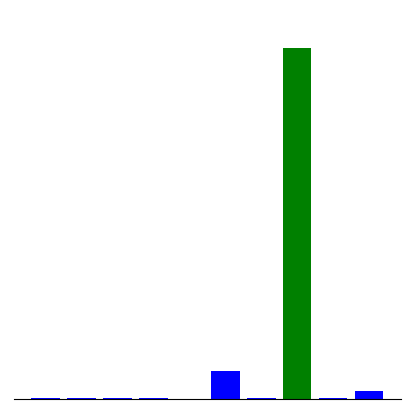} &

\includegraphics[width=0.055\textwidth]{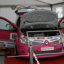} &
\includegraphics[width=0.055\textwidth]{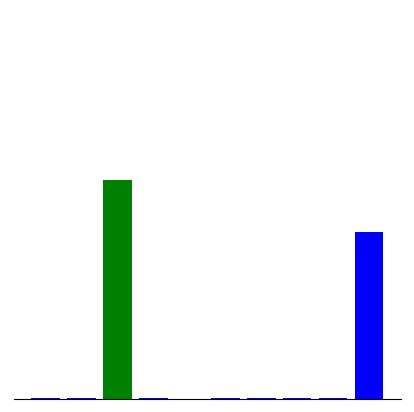} &

\includegraphics[width=0.055\textwidth]{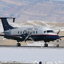} &
\includegraphics[width=0.055\textwidth]{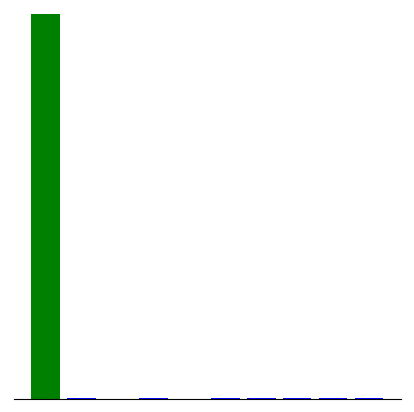} &

\includegraphics[width=0.055\textwidth]{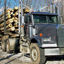} &
\includegraphics[width=0.055\textwidth]{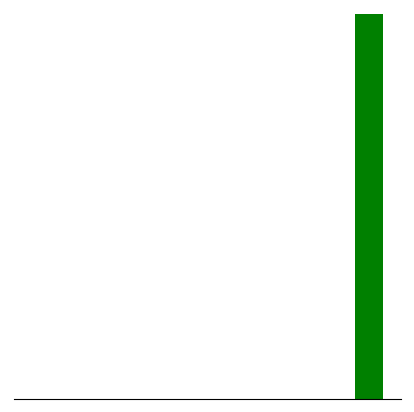} \\

\includegraphics[width=0.055\textwidth]{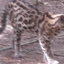} &
\includegraphics[width=0.055\textwidth]{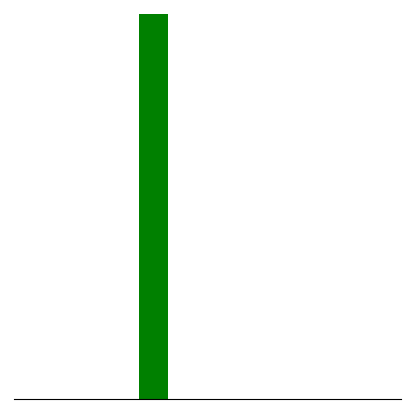} &

\includegraphics[width=0.055\textwidth]{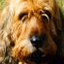} &
\includegraphics[width=0.055\textwidth]{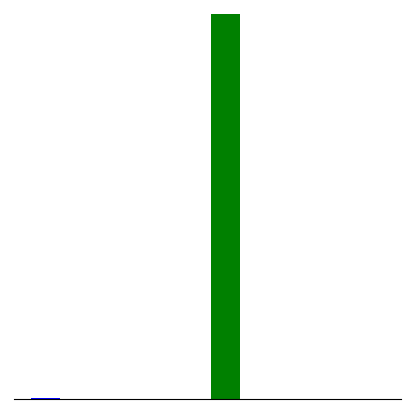} &

\includegraphics[width=0.055\textwidth]{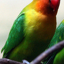} &
\includegraphics[width=0.055\textwidth]{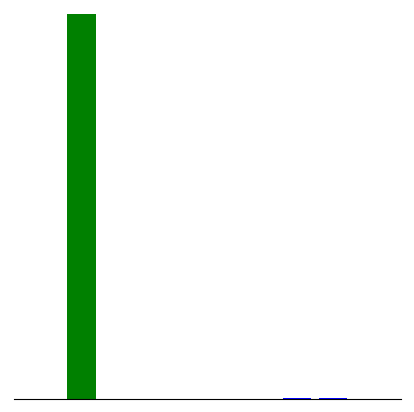} &

\includegraphics[width=0.055\textwidth]{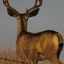} &
\includegraphics[width=0.055\textwidth]{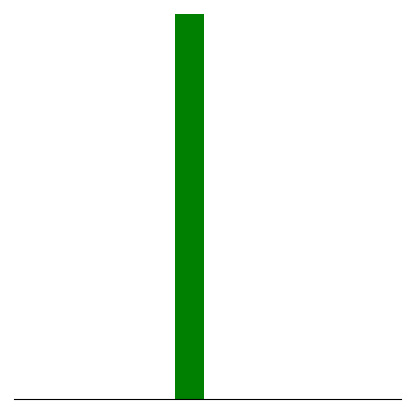} &

\includegraphics[width=0.055\textwidth]{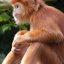} &
\includegraphics[width=0.055\textwidth]{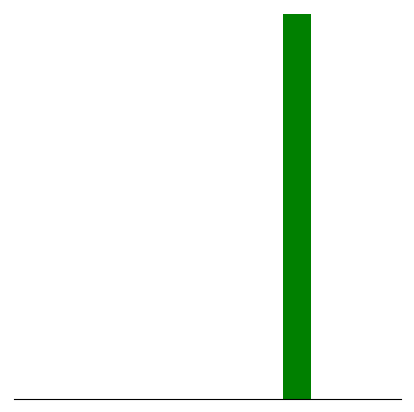} &

\includegraphics[width=0.055\textwidth]{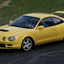} &
\includegraphics[width=0.055\textwidth]{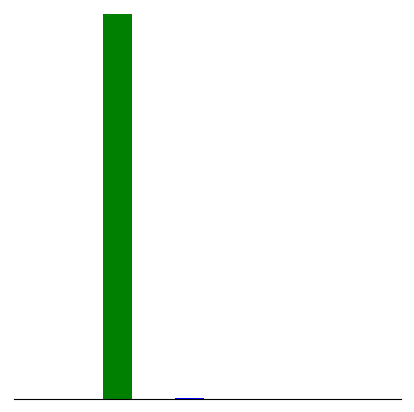} &

\includegraphics[width=0.055\textwidth]{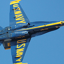} &
\includegraphics[width=0.055\textwidth]{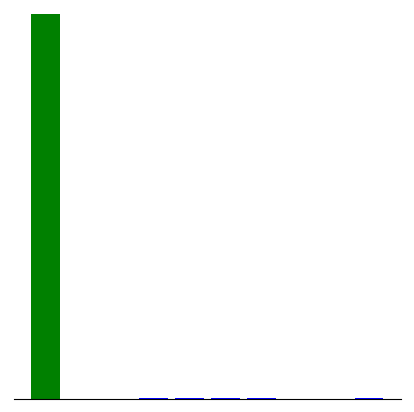} &

\includegraphics[width=0.055\textwidth]{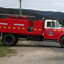} &
\includegraphics[width=0.055\textwidth]{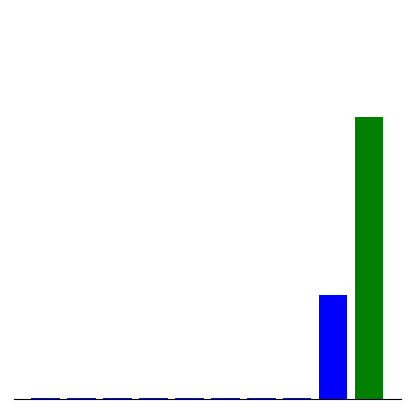} \\

\includegraphics[width=0.055\textwidth]{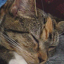} &
\includegraphics[width=0.055\textwidth]{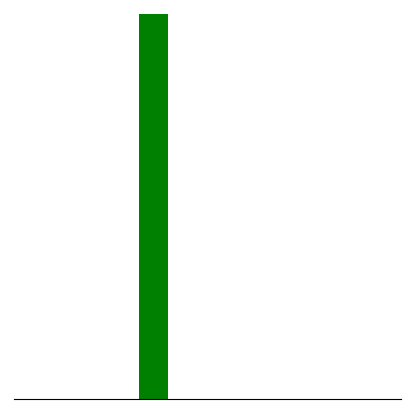} &

\includegraphics[width=0.055\textwidth]{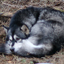} &
\includegraphics[width=0.055\textwidth]{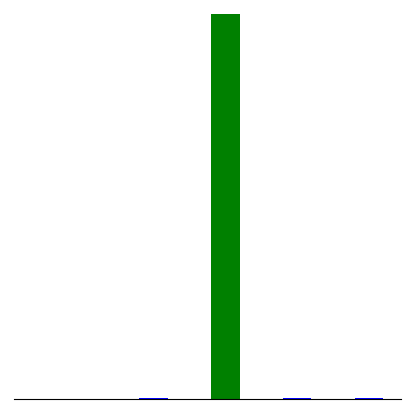} &

\includegraphics[width=0.055\textwidth]{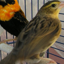} &
\includegraphics[width=0.055\textwidth]{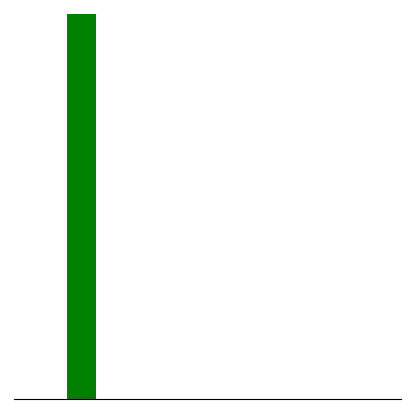} &

\includegraphics[width=0.055\textwidth]{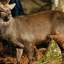} &
\includegraphics[width=0.055\textwidth]{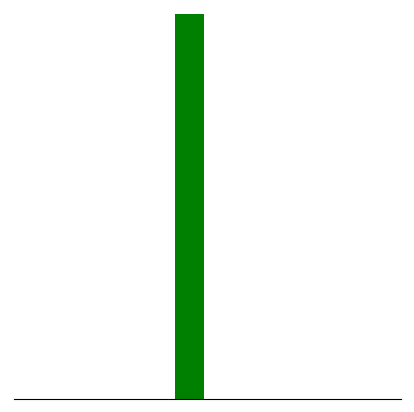} &

\includegraphics[width=0.055\textwidth]{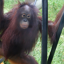} &
\includegraphics[width=0.055\textwidth]{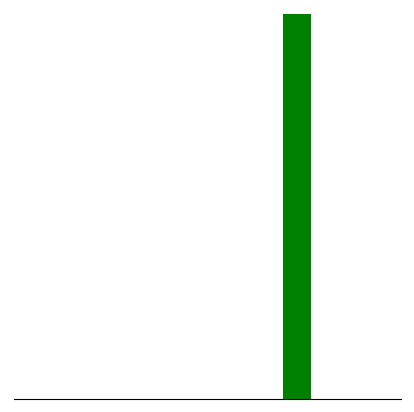} &

\includegraphics[width=0.055\textwidth]{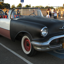} &
\includegraphics[width=0.055\textwidth]{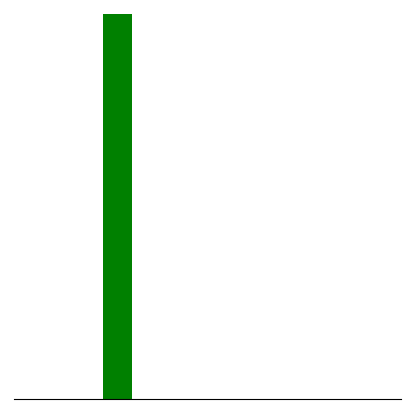} &

\includegraphics[width=0.055\textwidth]{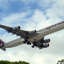} &
\includegraphics[width=0.055\textwidth]{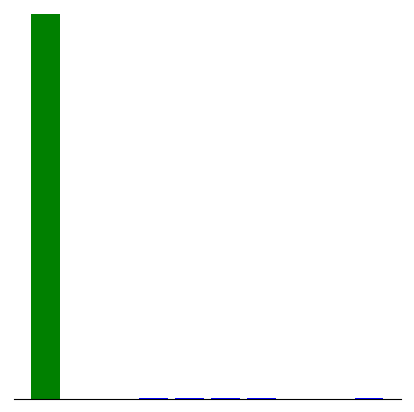} &

\includegraphics[width=0.055\textwidth]{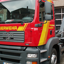} &
\includegraphics[width=0.055\textwidth]{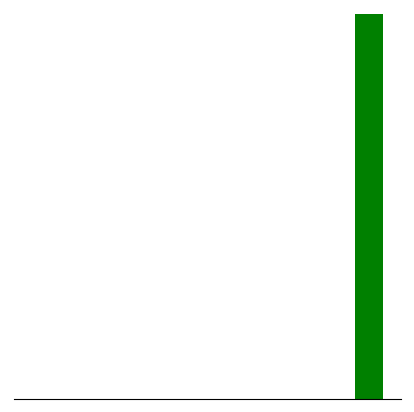} \\

\includegraphics[width=0.055\textwidth]{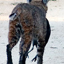} &
\includegraphics[width=0.055\textwidth]{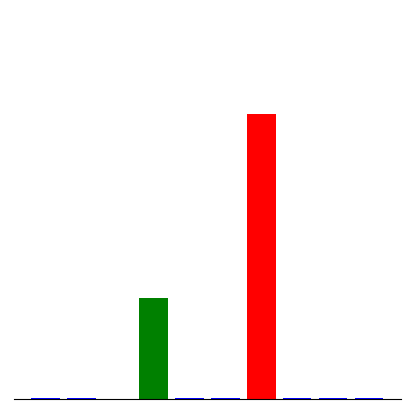} &

\includegraphics[width=0.055\textwidth]{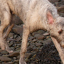} &
\includegraphics[width=0.055\textwidth]{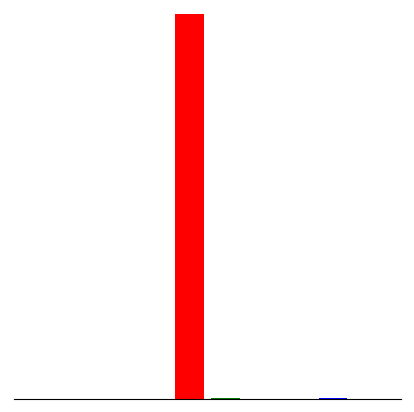} &

\includegraphics[width=0.055\textwidth]{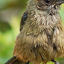} &
\includegraphics[width=0.055\textwidth]{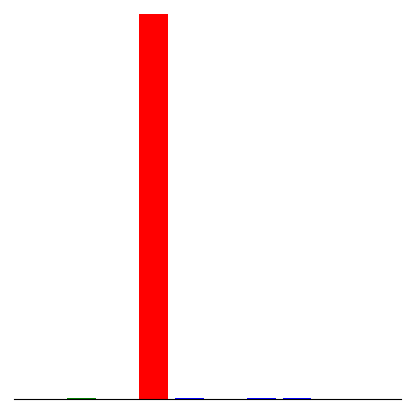} &

\includegraphics[width=0.055\textwidth]{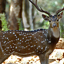} &
\includegraphics[width=0.055\textwidth]{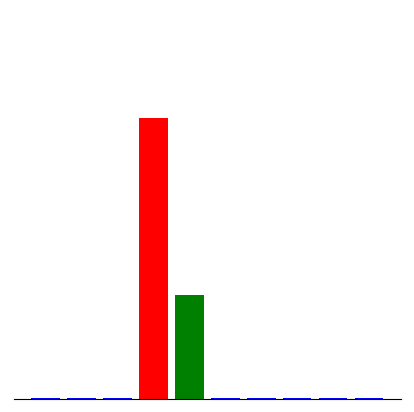} &

\includegraphics[width=0.055\textwidth]{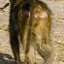} &
\includegraphics[width=0.055\textwidth]{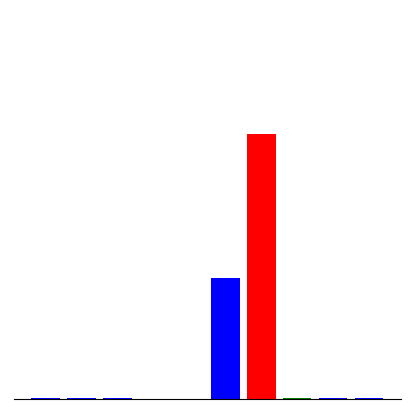} &

\includegraphics[width=0.055\textwidth]{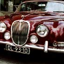} &
\includegraphics[width=0.055\textwidth]{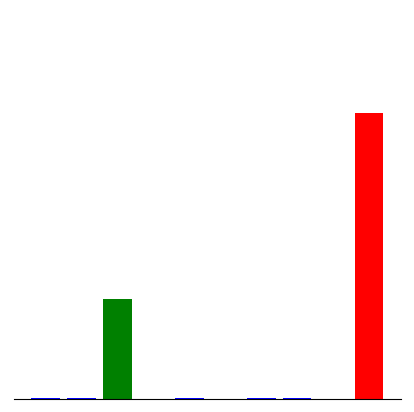} &

\includegraphics[width=0.055\textwidth]{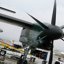} &
\includegraphics[width=0.055\textwidth]{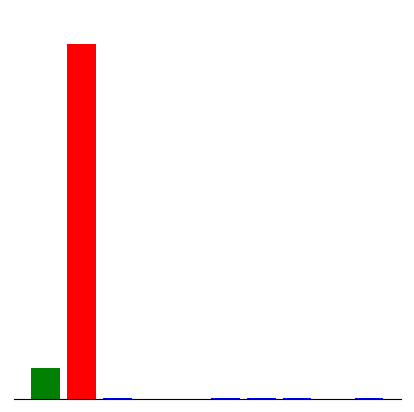} &

\includegraphics[width=0.055\textwidth]{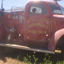} &
\includegraphics[width=0.055\textwidth]{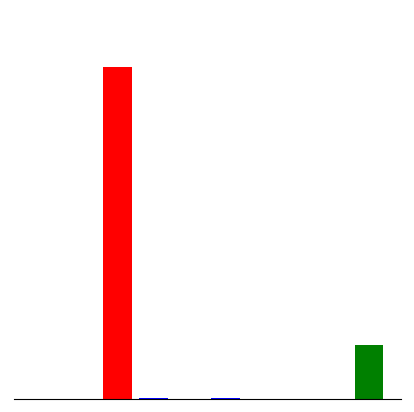} \\
\end{tabular}
\end{center}\vspace{-1em}
\caption{\textbf{Unsupervised image clustering (\methodnameshort) results on STL10.} Predicted cluster probabilities from the best performing head are shown as bars. Prediction corresponds to tallest, ground truth is green, incorrectly predicted classes are red, and all others are blue. The bottom row shows failure cases.}
\label{f:images_img_clus}
\end{figure*}
\begin{figure*}[t]
\TopFloatBoxes
\begin{floatrow}
\floatbox[\nocapbeside]{table}[0.27\textwidth]  
{
\caption{\textbf{Fully and semi-supervised classification.} Legend: *Fully supervised method. $\star$Our experiments with authors' code. $\dagger$Multi-fold evaluation.}
\label{t:iid_imgclus_semisup}
}
{
\scriptsize
\begin{tabular}{lc}
\toprule
& STL10 \\
\midrule
Dosovitskiy 2015~\cite{dosovitskiy2015discriminative}$\dagger$ & 74.2 \\
SWWAE 2015~\cite{zhao2015stacked}$\dagger$ & 74.3 \\
Dundar 2015~\cite{dundar2015convolutional}& 74.1 \\
Cutout* 2017~\cite{devries2017improved}& 87.3 \\
Oyallon* 2017~\cite{oyallon2017scaling}$\dagger$ & 76.0 \\
Oyallon* 2017~\cite{oyallon2017scaling}& 87.6 \\
DeepCluster 2018~\cite{caron2018deep} & 73.4$\star$ \cmt{428} \\
ADC 2018~\cite{haeusser2018associative} & 56.7$\star$ \\
DeepINFOMAX 2018~\cite{hjelm2018learning} & 77.0 \\
\methodnameshort plus finetune$\dagger$ & \textbf{79.2} \\
\methodnameshort plus finetune & \textbf{88.8} \cmt{650, 698} \\
\bottomrule
\end{tabular}}
\floatbox[\nocapbeside]{figure}[0.68\textwidth]  
{
\caption{\textbf{Semi-supervised overclustering.} Training with \methodnameshort loss to overcluster ($k>k_{gt}$) and using labels for evaluation mapping only. Performance is robust even with 90\%-75\% of labels discarded (left and center). STL10-$r$ denotes networks with output $k=\lceil1.4r \rceil$. Overall accuracy improves with the number of output clusters $k$ (right). For further details see supplementary material. }\label{f:imgclus_variation}
}
{
\includegraphics[width=0.215\textwidth,trim=0 0 0 1em, clip]{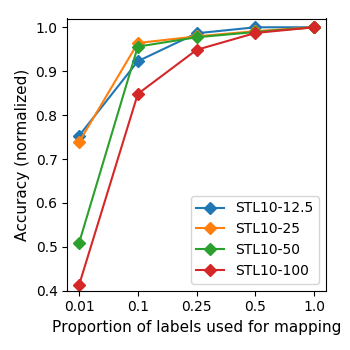}~~~%
\includegraphics[width=0.215\textwidth,trim=0 0 0 1em, clip]{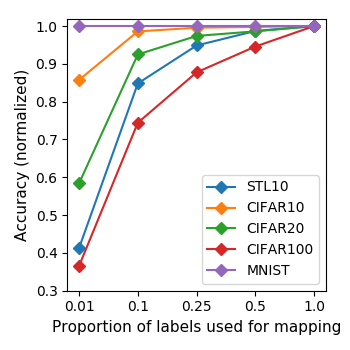}~~~%
\includegraphics[width=0.215\textwidth,trim=0 0 0 1em, clip]{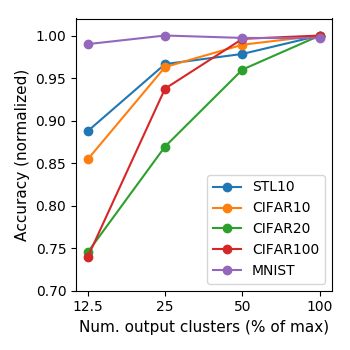}
}
\end{floatrow}
\end{figure*}

\paragraph{Datasets.}

We test on STL10, which is ImageNet adapted for unsupervised classification, as well as CIFAR10, CIFAR100-20 and MNIST. The main setting is pure unsupervised clustering (\methodnameshort) but we also test two semi-supervised settings: \emph{finetuning} and \emph{overclustering}.
For unsupervised clustering, following previous work~\cite{chang2017deep,xie2016unsupervised,yang2016joint}, we train on the full dataset and test on the labelled part; for the semi-supervised settings, train and test sets are separate.


As for DeepCluster~\cite{caron2018deep}, we found Sobel filtering to be beneficial, as it discourages clustering based on trivial cues such as colour and encourages using more meaningful cues such as shape.
Additionally, for data augmentation, we repeat images within each batch $r$ times; this means that multiple image pairs within a batch contain the same original image, each paired with a different transformation, which encourages greater distillation since there are more examples of which visual details to ignore~(\cref{s:equalization}).
We set $r\in[1,5]$ for all experiments. Images are rescaled and cropped for training (prior to applying transforms $g$, consisting of random additive and multiplicative colour transformations and horizontal flipping) and a single center crop is used at test time for all experiments except semi-supervised finetuning, where 10 crops are used.

\paragraph{Architecture.}
All networks are randomly initialised and consist of a ResNet or VGG11-like base $b$ (see sup.\ mat.), followed by one or more heads (linear predictors).
Let the number of ground truth clusters be $k_{gt}$ and the output channels of a head be $k$.
For \methodnameshort, there is a main output head with $k=k_{gt}$ and an auxiliary overclustering head~(\cref{f:overview}) with $k>k_{gt}$.
For semi-supervised overclustering there is one output head with $k>k_{gt}$.
For increased robustness, each head is duplicated $h=5$ times with a different random initialisation, and we call these concrete instantiations sub-heads.
Each sub-head takes features from $b$ and outputs a probability distribution for each batch element over the relevant number of clusters.
For semi-supervised finetuning~(\cref{t:iid_imgclus_semisup}), the base is copied from a semi-supervised overclustering network and combined with a single randomly initialised linear layer where $k=k_{gt}$.

\paragraph{Training.}
We use the Adam optimiser~\cite{kingma2014adam} with learning rate $10^{-4}$. For \methodnameshort, the main and auxiliary heads are trained by maximising ~\cref{e:loss_expanded} in alternate epochs.
For semi-supervised overclustering, the single head is trained by maximising~\cref{e:loss_expanded}. Semi-supervised finetuning uses a standard logistic loss.

\paragraph{Evaluation.}
We evaluate based on accuracy (true positives divided by sample size). For \methodnameshort we follow the standard protocol of finding the best one-to-one permutation mapping between learned and ground-truth clusters (from the main output head; auxiliary overclustering head is ignored) using linear assignment~\cite{kuhn2010hungarian}. While this step uses labels, it does not constitute learning as it merely makes the metric invariant to the order of the clusters.
For semi-supervised overclustering, each ground-truth cluster may correspond to the union of several predicted clusters.
Evaluation thus requires a many-to-one discrete map from $k$ to $k_{gt}$, since $k > k_{gt}$. This extracts some information from the labels and thus requires separated training and test set. Note this mapping is found using the training set (accuracy is computed on the test set) and does not affect the network parameters as it is used for evaluation only.
For semi-supervised finetuning, output channel order matches ground truth so no mapping is required.
Each sub-head is assessed independently; we report average and best sub-head (as chosen by lowest IIC loss) performance.

\paragraph{Unsupervised learning analysis.}
\methodnameshort is highly capable of discovering clusters in unlabelled data that accurately correspond to the underlying semantic classes, and outperforms all competing baselines at this task~(\cref{t:img_clus_iid}), with significant margins of $6.6\%$ and $9.5\%$ in the case of STL10 and CIFAR10. As mentioned in~\cref{s:related}, this underlines the advantages of end-to-end optimisation instead of using a fixed external procedure like k-means as with many baselines. The clusters found by \methodnameshort are highly discriminative~(\cref{f:images_img_clus}), although note some failure cases; as \methodnameshort distills purely visual correspondences within images, it can be confused by instances that combine classes, such as a deer with the coat pattern of a cat. Our ablations~(\cref{t:iid_imgclus_ablation}) illustrate the contributions of various implementation details, and in particular the accuracy gain from using auxiliary overclustering.

\paragraph{Semi-supervised learning analysis.}
For semi-supervised learning, we establish a new state-of-the-art on STL10 out of all reported methods by finetuning a network trained in an entirely unsupervised fashion with the \methodnameshort objective (recall labels in semi-supervised overclustering are used for evaluation and do not influence the network parameters). This explicitly validates the quality of our unsupervised learning method, as we beat even the supervised state-of-the-art~(\cref{t:iid_imgclus_semisup}). Given that the bulk of parameters within semi-supervised overclustering are trained
unsupervised (i.e. all network parameters), it is unsurprising that~\Cref{f:imgclus_variation} shows a 90\% drop in the number of available labels for STL10 (decreasing the amount of labelled data available from 5000 to 500 over 10 classes) barely impacts performance, costing just $\sim$10\% drop in accuracy. This setting has lower label requirements than finetuning because whereas the latter learns all network parameters, the former only needs to learn a discrete map between $k$ and $k_{gt}$, making it an important practical setting for applications with small amounts of labelled data.

\subsection{Segmentation}
\begin{figure*}
\vspace{-1.5em}

\includegraphics[width=\textwidth]{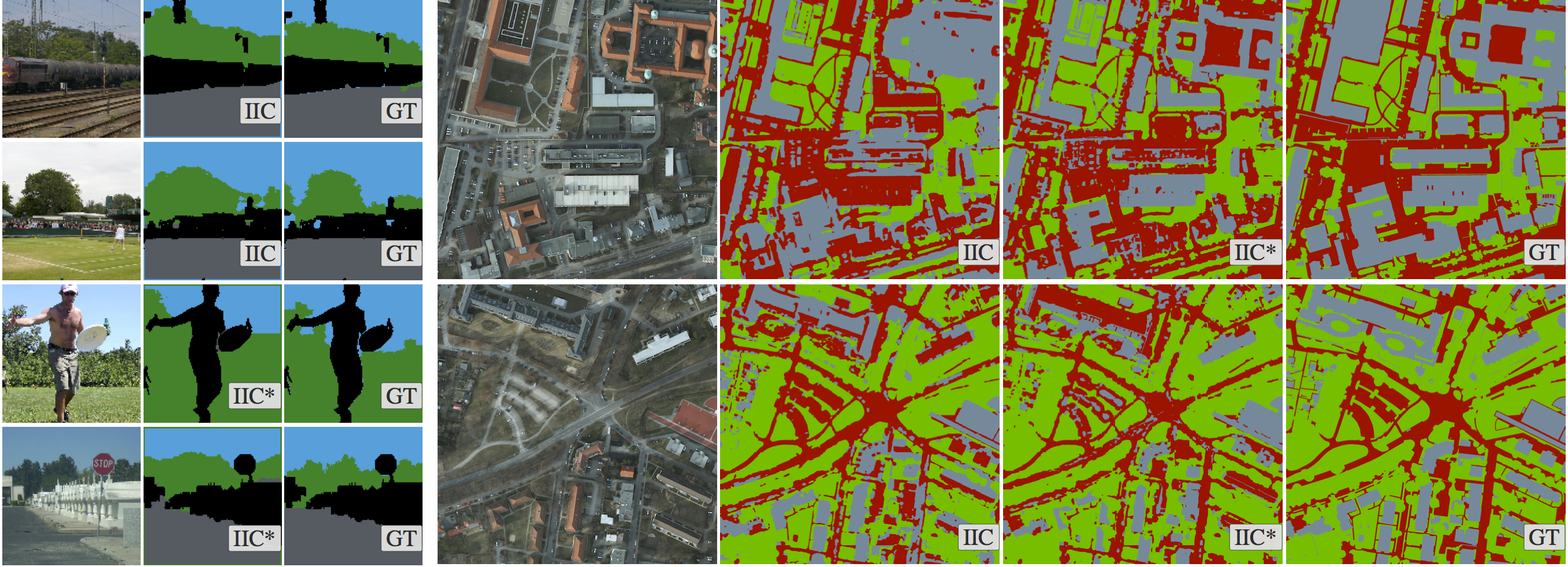}

\caption{\textbf{Example segmentation results (un- and semi-supervised).} Left: COCO-Stuff-3 (non-stuff pixels in black), right: Potsdam-3. Input images, IIC (fully unsupervised segmentation) and IIC* (semi-supervised overclustering) results are shown, together with the ground truth segmentation (GT).}
\label{f:images_img_seg}
\end{figure*}

\paragraph{Datasets.}
Large scale segmentation on real-world data using deep neural networks is extremely difficult without labels or heuristics, and has negligible precedent.
We establish new baselines on scene and satellite images to highlight performance on textural classes, where the assumption of spatially proximal invariance~(\cref{s:image_segmentation}) is most valid.
COCO-Stuff~\cite{caesar2016coco} is a challenging and diverse segmentation dataset containing ``stuff'' classes ranging from buildings to bodies of water.
We use the 15 coarse labels and 164k images variant, reduced to 52k by taking only images with at least 75\% stuff pixels.
COCO-Stuff-3 is a subset of COCO-Stuff with only sky, ground and plants labelled.
For both COCO datasets, input images are shrunk by two thirds and cropped to $128\times128$ pixels, Sobel preprocessing is applied for data augmentation, and predictions for non-stuff pixels are ignored.
Potsdam~\cite{potsdam} is divided into 8550 RGBIR $200\times200$ px satellite images, of which 3150 are unlabelled.
We test both the 6-label variant (roads and cars, vegetation and trees, buildings and clutter) and a 3-label variant (Potsdam-3) formed by merging each of the 3 pairs.
All segmentation training and testing sets have been released with our code.

\begin{table}
\setlength{\tabcolsep}{1pt}
\fontsize{8}{9}\selectfont 
\begin{tabular}{lcccc}
\toprule
& COCO-Stuff-3  & COCO-Stuff & Potsdam-3 & Potsdam  \\
\midrule
Random CNN & 37.3 \cmt{509} & 19.4 \cmt{512} & 38.2 \cmt{497} & 28.3 \cmt{500} \\
K-means~\cite{scikit-learn}$\dagger$ & 52.2 \cmt{527} & 14.1 \cmt{528} & 45.7 \cmt{501} & 35.3 \cmt{503} \\
SIFT~\cite{lowe2004distinctive}$\ddagger$& 38.1 \cmt{529} & 20.2 \cmt{530} & 38.2 \cmt{517} & 28.5 \cmt{518} \\
Doersch 2015~\cite{doersch2015unsupervised}$\ddagger$ & 47.5 \cmt{550} & 23.1 \cmt{551} & 49.6 \cmt{542} & 37.2 \cmt{546}\\
Isola 2016~\cite{isola2015learning}$\ddagger$ & 54.0 \cmt{534} & 24.3 \cmt{535} & 63.9 \cmt{532} & 44.9 \cmt{537}\\
DeepCluster 2018~\cite{caron2018deep}$\dagger$ $\ddagger$ & 41.6 \cmt{524} & 19.9 \cmt{553} & 41.7 \cmt{523} & 29.2 \cmt{525}\\
\midrule
\methodnameshort & \textbf{72.3} \cmt{555} & \textbf{27.7} \cmt{512} & \textbf{65.1} \cmt{545} & \textbf{45.4} \cmt{544} \\
\bottomrule
\end{tabular}
\caption{\textbf{Unsupervised segmentation.} \methodnameshort experiments use a single sub-head. Legend: $\dagger$Method based on k-means. $\ddagger$Method that does not directly learn a clustering function and requires further application of k-means to be used for image clustering.}\label{t:iid_seg}
\end{table}

\paragraph{Architecture.}
All networks are randomly initialised and consist of a base CNN $b$ (see sup. mat.) followed by head{}(s), which are $1\times1$ convolution layers.
Similar to~\cref{s:exp_img_clus}, overclustering uses $k$ 3-5 times higher than $k_{gt}$.
Since segmentation is much more expensive than image clustering (e.g.\ a single $200\times200$ Potsdam image contains 40,000 predictions), all segmentation experiments were run with $h = 1$ and $r = 1$ (sec.~\ref{s:exp_img_clus}).

\paragraph{Training.}
The convolutional implementation of \methodnameshort (\cref{e:info_seg}) was used with $d=10$. For Potsdam-3 and COCO-Stuff-3, the optional entropy coefficient~(\cref{s:equalization} and sup. mat.) was used and set to 1.5. Using the coefficient made slight improvements of 1.2\%-3.2\% on performance. These two datasets are balanced in nature with very large sample volume (e.g. $40,000 \times 75$ predictions per batch for Potsdam-3) resulting in stable and balanced batches, justifying prioritisation of equalisation. Other training details are the same as~\cref{s:exp_img_clus}.

\paragraph{Evaluation.}
Evaluation uses accuracy as in~\cref{s:exp_img_clus}, computed per-pixel.
For the baselines, the original authors' code was adapted from image clustering where available, and the architectures are shared with \methodnameshort for fairness. For baselines that required application of k-means to produce per-pixel predictions~(\cref{t:iid_seg}), k-means was trained with randomly sampled pixel features from the training set (10M for Potsdam, Potsdam-3; 50M for COCO-Stuff, COCO-Stuff-3) and tested on the full test set to obtain accuracy.

\paragraph{Analysis.}
 Without labels or heuristics to learn from, and given just the cluster cardinality (3), \methodnameshort automatically partitions COCO-Stuff-3 into clusters that are recognisable as sky, vegetation and ground, and learns to classify vegetation, roads and buildings for Potsdam-3~(\cref{f:images_img_seg}). The segmentations are notably intricate, capturing fine detail, but are at the same time locally consistent and coherent across all images. Since spatial smoothness is built into the loss~(\cref{s:image_segmentation}), all our results are able to use raw network outputs without post-processing (avoiding e.g. CRF smoothing~\cite{chen2018deeplab}). Quantitatively, we outperform all baselines~(\cref{t:iid_seg}), notably by $18.3\%$ in the case of COCO-Stuff-3. The efficient convolutional formulation of the loss~(\cref{e:info_seg}) allows us to optimise over all pixels in all batch images in parallel, converging in fewer epochs (passes of the dataset) without paying the price of reduced computational speed for dense sampling. This is in contrast to our baselines which, being not natively adapted for segmentation, required sampling a subset of pixels within each batch, resulting in increased loss volatility and training speeds that were up to 3.3$\times$ slower than \methodnameshort.

\section{Conclusions}\label{s:conc}
We have shown that it is possible to train neural networks into semantic clusterers without using labels or heuristics. The novel objective presented relies on statistical learning, by optimising mutual information between related pairs - a relationship that can be generated by random transforms - and naturally avoids degenerate solutions. The resulting models classify and segment images with state-of-the-art levels of semantic accuracy. Being not specific to vision, the method opens up many interesting research directions, including optimising information in datastreams over time. 

{\noindent\textbf{Acknowledgments.} We are grateful to ERC StG IDIU-638009 and EPSRC AIMS CDT for support.}
{\small\bibliographystyle{ieee_fullname}\bibliography{main}}
\end{document}